\documentclass[letterpaper, 10 pt, conference]{ieeeconf}  % Comment this line out if you need a4paper

\IEEEoverridecommandlockouts                              % This command is only needed if 
\usepackage{amssymb}  % assumes amsmath package installed

\usepackage{xcolor}
\usepackage{colortbl,booktabs}
\usepackage{graphicx}
\usepackage{xcolor}
\usepackage{fancyhdr}
\usepackage{tabularx}
\usepackage{multirow}

\title{\LARGE \bf
Multi-Slice Dense-Sparse Learning for Efficient Liver and Tumor Segmentation
}

\author{Ziyuan Zhao$^{1}$, Zeyu Ma$^{1, 2}$, Yanjie Liu$^{1, 2}$, Zeng Zeng$^{\dagger 1}$, Pierce KH Chow$^{3, 4}$
\thanks{$^{\dagger}$Corresponding author. This research is supported by Institute for Infocomm Research (I2R), Agency for Science, Technology and Research (A*STAR), Singapore. $^{1}$ Institute for Infocomm Research (I2R), Agency for Science, Technology and Research (A*STAR), Singapore. $^{2}$ National University of Singapore, Singapore. $^{3}$ Dept of Hepatopancreatobiliary Surgery and Transplant, National Cancer Center and Singapore General Hospital, Singapore. $^{4}$ Duke-NUS Medical School Singapore. This work was done when Zeyu and Yanjie were interns at I2R, A*STAR.}
}

\begin{document}

\maketitle
\thispagestyle{empty}
\pagestyle{empty}

%%%%%%%%%%%%%%%%%%%%%%%%%%%%%%%%%%%%%%%%%%%%%%%%%%%%%%%%%%%%%%%%%%%%
\thispagestyle{fancy}
\fancyfoot{}
\lfoot{\scriptsize{© 2021 IEEE.  Personal use of this material is permitted.  Permission from IEEE must be obtained for all other uses, in any current or future media, including reprinting/republishing this material for advertising or promotional purposes, creating new collective works, for resale or redistribution to servers or lists, or reuse of any copyrighted component of this work in other works.}}

% \thispagestyle{fancy}
% \fancyhead{}
% \lfoot{\scriptsize{© 2021 IEEE.  Personal use of this material is permitted.  Permission from IEEE must be obtained for all other uses, in any current or future media, including reprinting/republishing this material for advertising or promotional purposes, creating new collective works, for resale or redistribution to servers or lists, or reuse of any copyrighted component of this work in other works.}}

%%%%%%%%%%%%%%%%%%%%%%%%%%%%%%%%%%%%%%%%%%%%%%%%%%%%%%%%%%%%%%%%%%%%
\begin{abstract}

% Accurate automatic liver and tumor segmentation plays an important role in assisting treatment planing and disease monitoring. In recent years, deep convolutional neural network~(DCNNs), particularly U-Net, has achieved great success in 2D and 3D medical image segmentation. However, 2D convolutions cannot fully leverage the inter-slice information, while 3D networks is computationally expensive and memory intensive. In this work, we propose a 2.5D light-weight network based on nnU-net, in which, we use  depth-wise separable convolution instead of traditional convolutional layers. Extensive experiments and analysis on the LiTS dataset have demonstrated the superiority of the proposed method.

Accurate automatic liver and tumor segmentation plays a vital role in treatment planning and disease monitoring.
Recently, deep convolutional neural network~(DCNNs) has obtained tremendous success in 2D and 3D medical image segmentation.
However, 2D DCNNs cannot fully leverage the inter-slice information, while 3D DCNNs are computationally expensive and memory intensive.
To address these issues, we first propose a novel dense-sparse training flow from a data perspective, in which, densely adjacent slices and sparsely adjacent slices are extracted as inputs for regularizing DCNNs, thereby improving the model performance.
Moreover, we design a 2.5D light-weight nnU-Net from a network perspective, in which, depthwise separable convolutions are adopted to improve the efficiency.
Extensive experiments on the LiTS dataset have demonstrated the superiority of the proposed method.
\newline
\indent \textit{Clinical relevance}— The proposed method can effectively segment livers and tumors from CT scans with low complexity, which can be easily implemented into clinical practice.
\end{abstract}

%%%%%%%%%%%%%%%%%%%%%%%%%%%%%%%%%%%%%%%%%%%%%%%%%%%%%%%%%%%%%%%%%%%%%%%%%%%%%%%%
\section{INTRODUCTION}
Liver cancer is life-threatening and one of the most dangerous tumors to human health~\cite{Forner2015Hepatocellular}.
Computed tomography (CT) is one of the most effective non-invasive diagnostic imaging procedures to help doctors detect and characterize liver lesions~\cite{xu2020multi}.
Moreover, accurate localization and segmentation of liver and lesions is a crucial step for clinical diagnosis and surgical planning~\cite{gu2020multi}.
However, in routine clinical practice, manually segmenting liver and lesions from CT scans is time-consuming and error-prone.
Therefore, automatic computer-aided segmentation methods are urgently needed.

% According to WHO research~\cite{Forner2015Hepatocellular}, liver cancer is one of the five most threatening cancers to human health. Among the various imaging methods, CT~(Computerized Tomography) is the most common way to detect liver lesions. CT scans can be performed to assess the liver, including tumor volume, shape, location,~\emph{etc.}, to help doctors better judge tumor evaluation and treatment options. Since manual processing is time-consuming and error-prone, the automatic segmentation technology of liver and tumor will have a significant impact on disease monitoring. 

In recent years, deep learning has advanced the development of computer-aided diagnosis~\cite{hesamian2019deep}. In particular, fully convolutional networks (FCNs)~\cite{2015Fully,Ronneberger2015U}, \emph{e.g.}, 2D and 3D FCNs, have achieved promising performance for medical image segmentation~\cite{ben2016fully,christ2016automatic,pang2019modified,zheng2020automatic}.
2D FCNs have achieved good segmentation results in many medical imaging fields~\cite{gu2019net}, and are broadly implemented for liver and tumor segmentation in 2D CT slices~\cite{ben2016fully,christ2016automatic,chlebus2018automatic}. However, 2D FCNs ignore the inter-slice features in 3D volumetric CT scans, which limits the segmentation performance.
On the other hand, replacing 2D convolutions with 3D ones, 3D FCNs are capable to explore the inter-slice correlations and learn deep 3D representations with volumetric inputs, thereby obtaining more reliable results~\cite{huang2018robust}.
Regardless of the accuracy, high computational complexity and cost of 3D FCNs impede the broader clinical use. 
To solve this issue, much research effort has been devoted to maintaining an accuracy-complexity balance~\cite{2017H,zhang2019light,wardhana2021toward}, but most of them focus on constructing hybrid architectures using 2D and 3D convolutions together.

In this work, we first design a novel learning strategy from a data perspective, termed as~Dense-Sparse learning, in which, two types of adjacent slices,~\emph{i.e.}, densely adjacent slices and sparsely adjacent slice, are feed into 2D FCNs to probe the inter-slice information with different strides.
Furthermore, traditional convolutions in 2D nnU-Net~\cite{2018nnU} are replaced with depthwise separable convolutions from the perspective of network to form a 2.5D light-weight nnU-Net for improving the efficiency.
We extensively evaluate the proposed method on the LiTS dataset~\cite{bilic2019liver}. Experimental results demonstrate that the proposed method can achieve comparable performance on liver and tumor segmentation with much fewer parameters than 3D nnU-Net.

% Moreover, we design a 2.5D light-weight nnU-Net from a network perspective, in which, depthwise separable convolutions are adopted to improve the efficiency.
% Extensive experiments and analysis on the LiTS dataset have demonstrated the superiority of the proposed method.

% To solve the above problems, we present a 2.5D light-weight network based on nnU-net~\cite{2018nnU} with a novel sampling method Dense-Sparse sampling. The 2.5D network can improve the deficiencies of both 2D network and 3D network with the input of adjacent slices from the entire slice on 2D network architecture~\cite{vorontsov2018liver}. In training progress, we use a two-stage curriculum~\cite{li2020new} substitute for the cascade structure and Dense-Sparse sampling instead of traditional sampling. Extracting both densely adjacent slices and sparsely adjacent slices as inputs can increase the diversity of network inputs to prevent overfitting. Moreover, we use depth-wise separable convolution to replace the traditional convolutional layers which can make the network structure lighter. The result of the experiment shows the proposed method achieves comparable performance to the baseline method 3D nnUNet on the LiTS dataset~\cite{bilic2019liver}.

\begin{figure*}[htb]
\centering
\includegraphics[width=0.95\textwidth]{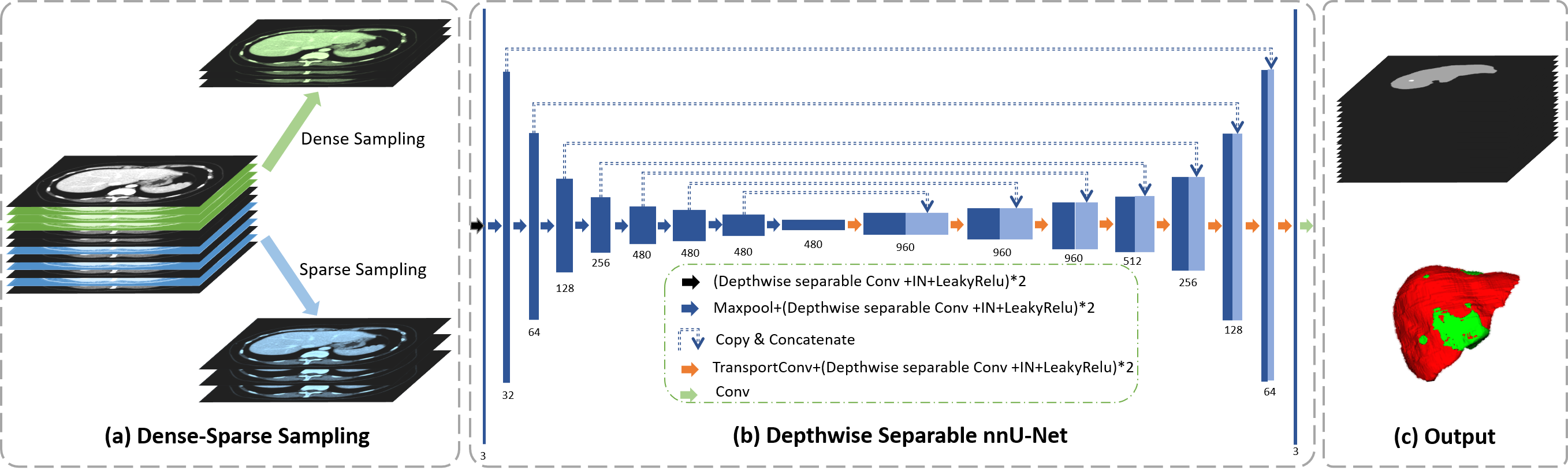}
\caption{Overall pipeline of our method: (a) Dense-Sparse Sampling: densely adjacent slices and sparsely adjacent slices with different stride $S$ are cropped from volumetric images for DCNNs. (b) Depthwise Separable nnU-Net: Depthwise separable convolutions are performed on nnU-Net for each input channel independently to reduce the parameters (c) Output: the output 2D segmentation results are stacked for generating 3D segmentation volumes.}
\label{fig:pipeline}
\end{figure*}

\section{RELATED WORK}

In the past decades, deep learning has received much attention on various computer vision tasks, such as classification and segmentation~\cite{litjens2017survey}. A lot of methods based on deep convolutional neural network~(DCNNs) have been proposed for liver and tumor segmentation~\cite{ben2016fully,chlebus2018automatic,vorontsov2018liver, Ronneberger2015U}.
Ben-Cohen~\emph{et al.}~\cite{ben2016fully} present an fully convolutional network~(FCN) for liver and tumor segmentation.
Chlebus~\emph{et al.}~\cite{chlebus2018automatic} propose a 2D U-Net with object-based post-processing, obtaining high performance in liver and tumor segmentation.
Vorontsov~\emph{et al.}~\cite{vorontsov2018liver} employ two parallel U-Nets~\cite{Ronneberger2015U} for joint liver and tumor segmentation.
These methods take 2D slices as inputs and thus ignore the contextual information between slices, limiting the segmentation performance, while 3D FCNs~\cite{dou20163d,lu2017automatic,hu2016automatic} are supposed to consider the inter-slice information, obtaining better segmentation results than 2D models.
For instance, Dou~\emph{et al.}~\cite{dou20163d} introduce a deep supervision mechanism into a 3D FCN for boosting the segmentation performance.
Nevertheless, training such 3D FCNs is time-consuming and memory-consuming, which limits the wide applications of 3D models.

To probe the inter-slice information while reducing the computational complexity, many methods from different perspectives have been proposed, including 2.5D models~\cite{han2017automatic,wardhana2021toward} and hybrid 2D–3D models~\cite{2017H,zhang2019light}. 
For example, in~\cite{han2017automatic}, adjacent slices from 3D CT scans are used to train a 2.5D FCN, while Li~\emph{et al.}~\cite{2017H} design a hybrid densely connected UNet~(H-DenseUNet) to jointly explore intra-slice and inter-slice features.
These methods alleviate the problems of 2D and 3D models to a certain extent. Differently, we consider both data and network perspectives in our framework for more efficient liver and tumor segmentation.

\section{METHODOLOGY}

% Our 2.5D light-weight network architecture is plotted in Fig.~\ref{fig:pipeline}, which can be summarized as follows. We use nnU-Net ~\cite{2018nnU}  to extract the LiTS dataset fingerprint (such as image sizes, voxel spacings, intensity information \emph{etc.}), create the training plan for different network architecture(including 2D U-Net, 3D U-Net,3D U-Net cascade). Here we select 2D training plan, including cropping, resampling and normalization, to preprocess the LiTS data. After data augmentation, such as elastic deformation, crop and rotation, our training process will be separated into two-stage. The first stage uses Dense-Sparse sampling. For every slice, it has half probability of using sparse sampling, else it will use dense sampling. Then, both the second stage and the inference stage will use fully-dense sampling.
% The proposed approach achieves promising performance without ensembling or post-processing and demonstrates superiority over the nnU-Net baseline.

The proposed framework is presented in Fig.~\ref{fig:pipeline}. Firstly, Dense-Sparse (DS) sampling strategy is used to generate two different inputs,~\emph{i.e.}, densely adjacent slices and sparsely adjacent slices, for improving the model performance. Then, a 2.5D nnU-Net with depthwise separable convolutions~(DS nnU-Net) is implemented as the segmentation model, in which, all convolutional layers are replaced with depthwise separable convolutions to reduce the computational complexity. We further design a two-stage dense-sparse-dense (DSD) training strategy, in which, densely adjacent slices and sparse adjacent slices are randomly selected to train the network at the first stage for fast convergence, while only densely adjacent slices are fed into the network at the second stage to improve the model robustness.

\subsection{Dense-Sparse Sampling}

Let $\mathbf{I} \in R^{N \times W \times H \times T}$ denote the training samples of height $H$, width $W$ and thickness $T$, where $N$ is the batch size. For a 2D network, only one slice ($T = 1$) is used to generate the segmentation mask, lacking the context information for volumetric medical image segmentation. In contrast, a 3D network takes the whole volumes (all $T$ slices) as inputs, suffering from high computational costs. In common, a 2.5D network uses a stack of $T$ continuous slices during training and generates the segmentation mask for the central slice at the inference stage. For instance, if $T = 3$, the densely adjacent slices ${1,2,3}$ would be the input of the 2.5D network to produce a 2D mask for slice $2$. The compromise can avoid high GPU memory consumption of 3D convolutions while providing inter-slice information. Beyond the densely adjacent slices, we propose a novel dense-sparse sampling method to generate densely adjacent slices and sparsely adjacent slices, as shown in Fig.~\ref{fig:pipeline}~(a).

For the sake of simplicity, we take $T = 3$ as an illustration. Let $S_i$ be the slice indexed by $i$ in a CT scan $|V|$, where $i=\{1,2,3, \ldots,|V|\}$. In dense-sparse (DS) sampling, $\mathbf{I}_{ds}$ is a 3D input of thickness $T = 3$, which is described as:

\begin{equation}
\mathbf{I}_{ds}=\{S_{i-(T // 2+s-1)}\} \cup\{S_{i}\} \cup\{S_{i+(T // 2+s-1)}\}, 
\end{equation}
where $//$ is integer division, and $s$ is the stride of sampling. In the case of edge slices, slices extending beyond the volume were repeated. DS sampling shares the similar spirit of the convolution kernels, in which, the stride in two dimensions is for the height and the width movement, while the stride of DS sampling is for the movement along the dimension of thickness. More specifically, when $s = 1$, densely adjacent slices $\mathbf{I}_{dense}$ will be generated, while $s > 1$ is for dense sampling to produce sparsely adjacent slices $\mathbf{I}_{sparse}$. With $\mathbf{I}_{dense}$ and $\mathbf{I}_{sparse}$ together, the network can learn inter-slice features efficiently. In our experiments, we set $s = 1$ and $s = 2$ for dense sampling and sparse sampling, respectively.

\subsection{Depthwise Separable nnU-Net}

nnU-Net~\cite{2018nnU} is a self-adapting framework based on generic U-Net architectures, such as 2D and 3D U-Net. With various kinds of data augmentation, nnU-Net makes full use of the potentials of U-Net. Considering the memory constraints and negative impact from cropped patches for 3D network training, we select 2D nnU-Net as our backbone and design a depthwise separable nnU-Net (DS nnU-Net), which employ depthwise separable convolutions instead of standard convolution layers to further ease the computational burden, as shown in Fig.~\ref{fig:pipeline}~(b).

Let us consider a convolutional layer with an input feature map 
$\mathbf{F}^I \in \mathbf{R}^{W_I\times H_I\times C_I}$ 
and output feature map $\mathbf{F}^O \in \mathbf{R}^{W_O\times H_O\times C_O}$, where $W_{*}$, $H_{*}$ and $C_{*}$ are spatial width, spatial height and the number of channels, respectively. The output of a standard convolution layer with kernel
$\mathbf{K}^{S}  \in \mathbf{}{R}^{X\times Y\times C_I \times C_O}$ can be formatted as: 

\begin{equation}
S C\left(\mathbf{K}^{S}, \mathbf{F}^I\right)_{w, h, c_O}=\sum_{x, y, c_I}^{X, Y, C_I} \mathbf{K}_{x, y,c_I, c_O}^{S} \cdot \mathbf{F}^I_{w, h, c_I}.
\end{equation}

A depthwise separable (DS) convolution can be split into 2 separate kernels, \emph{i.e.}, the depthwise convolution and the pointwise convolution. In depthwise convolution, each kernel $\mathbf{K}^{D} \in \mathbb{R}^{X\times Y\times C_I}$ iterates one channel of the input feature map $\mathbf{F}^I$. The output the depthwise convolution of can be computed as

\begin{equation}
D C\left(\mathbf{K}^{D}, \mathbf{F}^I\right)_{w, h, c_O}=\sum_{x, y, c_I}^{X, Y, C_I} \mathbf{K}_{x, y,c_I}^{D} \cdot \mathbf{F}^I_{w, h, c_I}.
\end{equation}

Following that, the pointwise convolution $K^P  \in \mathbb{R}^{1 \times 1 \times C_I \times C_O}$, also known as $1 \times 1 $ convolution, is applied to increase the number of channels. The mathematical formulation is

\begin{equation}
P C\left(\mathbf{K}^{P}, \mathbf{F}^I\right)_{w, h, c_O}=\sum_{x, y, c_I}^{X, Y, C_I} \mathbf{K}_{c_I, c_O}^{P} \cdot \mathbf{F}^I_{w, h, c_I}.
\end{equation}

If we have a $3 \times 3$ convolution with input channel $c$ and output channel $c$, for standard convolution, it has $9c^2$ parameters, while the depthwise separable convolution contains $9c + c^2$ parameters, which approximately reduce the computational complexity by a factor of $c$. We replace all standard convolutions with DS convolutions in our DS nnU-Net, which has only $7.7$ million (M) parameters, while 2D nnU-Net has more than $40$ M parameters.

% After applying depth-wise convolution block, our U-Net has $7,761,406$ parameters in total, while unmodified U-Net has more than $30,000,000$ previously.

% It is difficult to train a CNN with multiple branches, especially for multiple views. 
% There are no direct weights shared among the layers, which may cause negative knowledge transfer across multiple views. 
% We expect a smooth and fast convergence for robust results. To this end, we propose a network initialization scheme and a progressive learning strategy to facilitate model optimization, which is described as follows.
% densely adjacent slices and sparse adjacent slices are randomly selected to train the network at the first stage for fast convergence, while only densely slices are fed into the network for the second stage to improve the model robustness.
\subsection{Dense-Sparse-Dense Training}
Since densely adjacent slices and sparsely adjacent slices have different views of the data, which may cause negative transfer across views. To this end, we proposed a two-step progressive learning strategy, namely Dense-Sparse-Dense (DSD) training to facilitate model optimization. In the first $DS$ (Dense-Sparse) step, we randomly input densely adjacent slices and sparsely adjacent slices to train and regularize the network for fast convergence. In the second $D$ (Dense) step, we retrain the network with all densely adjacent slices for increasing the model capacity without overfitting.

\section{EXPERIMENTS}
\subsection{Dataset and Experimental Settings}
We evaluate the proposed method on the Liver Tumor Segmentation (LiTS) dataset~\cite{bilic2019liver}, which includes $201$ CT scans ($131$ for training and $70$ for testing). Since the ground truths of testing data are not publicly available, for a fair comparison, we randomly select $105$ volumes for training and the remaining $26$ for testing in our experiments.  The CT volumes are resized to multiple slices of size $512 \times 512$ after resampling and normalization. Following the settings in nnU-Net, we train the network with the combination of cross-entropy loss and dice loss. We implement DS sampling with thickness $T = 7$. For DSD training, we set $400$ and $600$ epochs for $DS$ step and $D$ step, respectively. 

According to the evaluation procedures of the LiTS challenge, we evaluated the liver and tumor segmentation performance using the only golden indicator, Dice per case score, which refers to an average of Dice per volume score.

We compare our method with 2D nnU-Net and 3D nnU-Net with full resolution. Besides, to validate the effectiveness of different components of our pipeline, the following variants are evaluated.
\begin{itemize}
    \item nnU-Net-DS: 2D nnU-Net with the proposed dense-sparse sampling.
    \item nnU-Net-DSD: 2D nnU-Net with the proposed dense-sparse-dense training strategy.
    \item DS nnU-Net-DSD: Depthwise Separable nnU-Net with the proposed dense-sparse-dense training strategy.
\end{itemize}

\begin{table}[thb]
\centering
\caption{Segmentation results of different methods on LiTS dataset }
\label{mytable_model}
\begin{tabular}{c|c|c|c} 
\hline
\multirow{2}{*}{Method} & Lesion        & Liver                               & \multirow{2}{*}{Params (M)}  \\ 
\cline{2-3}
                        & \multicolumn{2}{c|}{Dice per case (mean (std), \%)} &                                      \\ 
\hline
2D nnU-Net              & 0.801 (0.024) & 0.960 (0.004)                       & 41                                   \\
3D nnU-Net              & 0.827 (0.055) & 0.965 (0.004)                       & 37                                   \\ 
\hline
nnU-Net-DS             & 0.804 (0.041) & 0.962 (0.002)                       & 41                                   \\
nnU-Net-DSD             & 0.815 (0.032) & 0.963 (0.003)                       & 41                                   \\ 
\hline
DS nnU-Net-DSD          & 0.814 (0.025) & 0.962 (0.003)                       & 7                                    \\
\hline
\end{tabular}
\end{table}

\subsection{Results and Discussions}

The experimental results are shown in Table~\ref{mytable_model}. For a fair comparison, we train all models with $1000$ epochs. Apparently, 3D nnU-Net outperforms 2D nnU-Net on liver and tumor segmentation because it randomly samples 3D patches from CT volumes, which is capable to capture 3D contextual information for improved performance. 3D nnU-Net has fewer layers and parameters than 2D nnU-Net, but more training and inference time is required. Although 3D networks can effectively improve segmentation performance, it is also important to pay attention to model training. It is well noted that the proposed dense-sparse sampling can help improve the performance of 2D nnU-Net on liver and tumor segmentation. With DS sampling and DSD training strategy together, 2D nnU-Net achieved comparable performance with 3D nnU-Net. The results have demonstrated the effectiveness of the proposed DS sampling and DSD training strategy for improving segmentation performance without carefully modified architectures. On the other hand, we implemented DS nnU-Net with the proposed sampling and training strategies. It is observed that there is no significant performance degradation on liver and tumor segmentation with much fewer parameters, which is around $1/6$ of 2D nnU-Net. Furthermore, the training time is shortened by around $25\%$ on a single RTX 2020ti GPU.

\begin{figure}[!thb]
\centering
\includegraphics[width=0.4\textwidth]{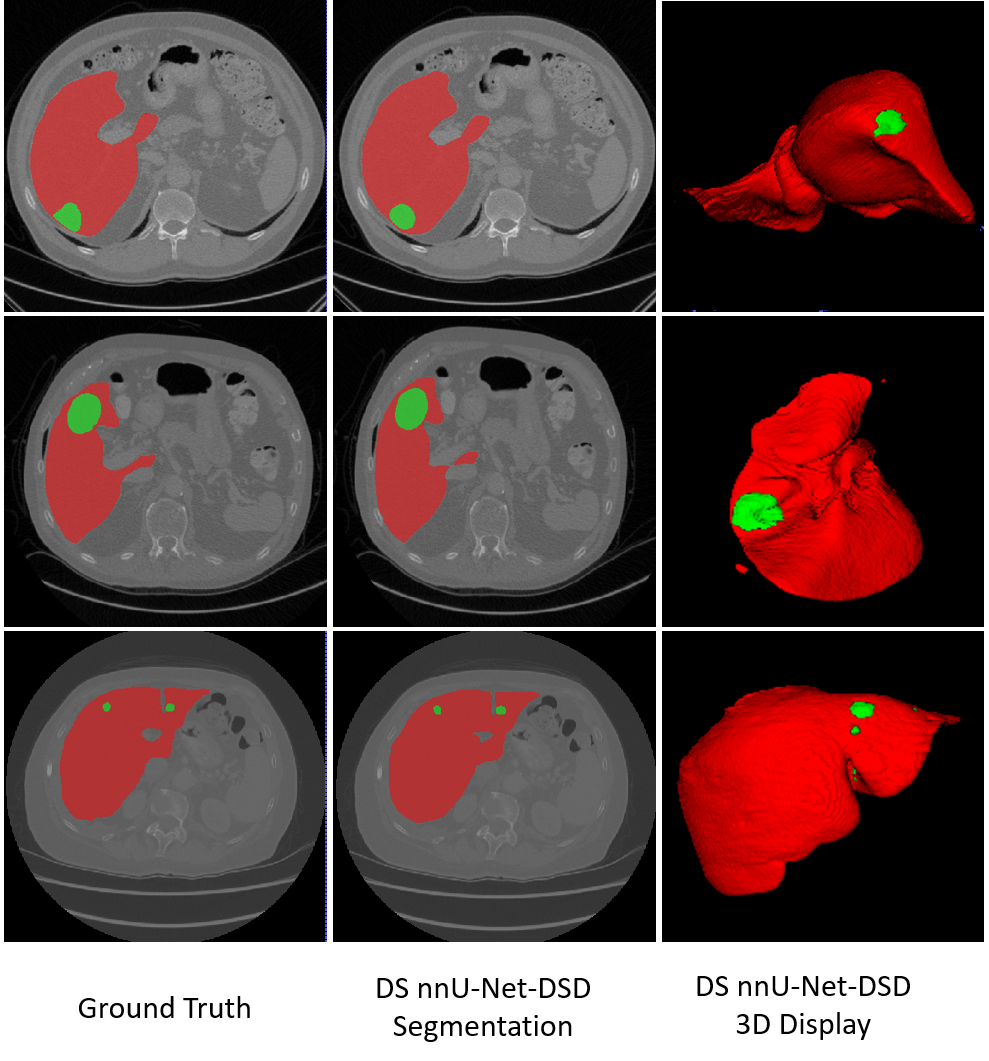}
\caption{Examples of segmentation result and 3D display by the proposed method (DS nnU-Net-DSD) on LiTS dataset.}
\label{fig:visua}
\end{figure}

Fig.~\ref{fig:visua}  shows the visualization results of our method DS nnU-Net-DSD. We can see that the masks of our method are close to the ground truth labels, which further shows the feasibility of the proposed method for efficient liver and lesion segmentation.

% Both the 3D model and the 2.5D model have better performance than the 2D model. Although 3D U-net depth and width are limited by GPU memory, 3D contextual features still have potential representation capability, this can also be proved by the effect of dense sampling model is better than 2D nnU-Net.

%  Compared with the two-stage training model, single-stage training strategy, including dense sampling, or dense-sparse sampling, has weaker performance. We believe that Dense-Sparse sampling not only improves the receptive field of the neurons, it can also learn the inter-slice information of the CT images, which is like we use different spacing to training. Not only that, our network robustness and generalization have also been significantly improved. For dense and sparse sampling models, adding a stage that uses single dense sampling can improve the stability of the model's output.
% Compared with ordinary convolutional networks, our networks with deep-wise separable convolution block has similar score. This expresses that reducing the number of parameters, even to three-quarters, does not reduce the accuracy of model. Training speed of our model has also be greatly increased. Take dense sampling as an example, depth separable convolution block can shorten training time by 25$\%$ on RTX 2080ti.

\section{CONCLUSIONS}

In this work, we design a novel end-to-end deep learning framework from both perspectives of data and network for liver and tumor segmentation. Extensive experiments show that the proposed approach can obtain accurate segmentation results, as well as speed up the training and inference process with only $7$ M parameters. Moreover, ablation studies demonstrate the effectiveness of different components in our framework. In the future, we will implement the proposed method in other medical image segmentation tasks to evaluate the generalization capability.

% In this paper, we have presented a 2.5D light-weight network with two-stage training strategy for liver and tumor segmentation. Compared with the commonly used 2D or 2.5D structure, the proposed approach can obtain more precise segmentation and speed up training and inference time. On the LiTS dataset, our novel sampling method, Dense-Sparse sampling, achieves promising performance, which demonstrates the superiority over other sampling methods. Moreover, the comparison among different stage models further justifies the reasonability of our design.

% In future research, we can implement more residual or attention block, and validate the proposed method on other popular medical image segmentation benchmarks.

\bibliographystyle{IEEEbib}
\bibliography{refs1.bib}

\end{document}